\documentclass[letterpaper,conference,final]{IEEEtran}

\usepackage{cite}
\usepackage[pdftex]{graphicx}
\usepackage[cmex10]{amsmath}\interdisplaylinepenalty=2500
\usepackage{url}
\usepackage{mathrsfs,amsfonts}

\newcommand*{\permcomb}[4][0mu]{\mkern#1#2^{#3}_{#4}}

\newcommand*{\comb}[1][0mu]{\permcomb[#1]{C}}

\DeclareMathOperator*{\argmin}{arg\,min}

\usepackage[noend]{algorithmic}
\usepackage{algorithm}
\renewcommand{\algorithmicrequire}{\textbf{Input:}}

\algsetup{indent=0.5cm}

\begin{document}

\title{A Genetic Algorithm to Discover Flexible Motifs with Support}

\author{\IEEEauthorblockN{
	Joan Serr\`a\IEEEauthorrefmark{1},
	Aleksandar Matic\IEEEauthorrefmark{1},
	Josep Lluis Arcos\IEEEauthorrefmark{2}, and
	Alexandros Karatzoglou\IEEEauthorrefmark{1}
}
	\IEEEauthorblockA{\IEEEauthorrefmark{1}Telef\'onica Research, Barcelona, Spain. %\\ 
	Email: firstname.lastname@telefonica.com 
	}
	\IEEEauthorblockA{\IEEEauthorrefmark{2}IIIA-CSIC, Bellaterra, Spain. %\\ 
	Email: arcos@iiia.csic.es 
	}
}

\maketitle

\begin{abstract}
Finding repeated patterns or motifs in a time series is an important unsupervised task that has still a number of open issues, starting by the definition of motif. In this paper, we revise the notion of motif support, characterizing it as the number of patterns or repetitions that define a motif. We then propose GENMOTIF, a genetic algorithm to discover motifs with support which, at the same time, is flexible enough to accommodate other motif specifications and task characteristics. GENMOTIF is an anytime algorithm that easily adapts to many situations: searching in a range of segment lengths, applying uniform scaling, dealing with multiple dimensions, using different similarity and grouping criteria, etc. GENMOTIF is also parameter-friendly: it has only two intuitive parameters which, if set within reasonable bounds, do not substantially affect its performance. 
We demonstrate the value of our approach in a number of synthetic and real-world settings, considering traffic volume measurements, accelerometer signals, and telephone call records.
%We demonstrate the value of our approach in a number of synthetic and real-world settings, including traffic volume measurements and accelerometer signals.
\end{abstract}

\IEEEpeerreviewmaketitle

\section{INTRODUCTION}

Discovering repeated short-length segments, commonly called motifs, in a time series is a fundamental task that has a number of important applications. Beyond data exploration and assessment, motif discovery is often a key component of many higher-level algorithms~\cite{Mueen14WIRES}, both of supervised and unsupervised nature. Due to this importance, a myriad of approaches have been proposed. Today, these can be categorized from different perspectives or dimensions, starting by the definition of a motif.

There are four main aspects that define a time series motif~\cite{Mueen14WIRES}: (a) the relative position of the segments, (b) the similarity between segments, (c) the length of the segments, and (d) the support of the motif. To determine~(a), pioneering approaches used approximate representations of the time series, defined a fixed segment length, and chose a static threshold to distinguish between similar and dissimilar segments~\cite{Chiu03KDD,Lin02WTDM}. Such approximate approaches have proven to be suitable in a number of applications, thus the work in this domain has continued until today~\cite{Begum14VLDB,Castro10SDM,Li12SDM}. However, so-called exact algorithms were recently introduced~\cite{Mueen09SDM} which, while being as efficient as some approximate ones, deal with the raw time series and do not require a similarity threshold. More recently, further developments have allowed to work with a range of segment lengths~\cite{Mueen13ICDM,Yingcharxxx13ICDM}, thus reducing the constraint of specifying a fixed value for such a crucial parameter.

Out of the four previous aspects, the introduction of the notion of motif support (d) remains the most elusive one. The current definitions of exact-based motifs are typically rooted on segment pairs~\cite{Mueen09SDM}. Yet there is no specific reason for constraining the definition of a motif to segment pairs. The alternative definition may be based on groups of segments, where the number of segments $s$ represents the support of the motif. This way, a definition based on segment pairs becomes just a special case in which $s=2$. Conceptually, this notion of motif support starts bridging the gap between motif discovery and subsequence time series clustering~\cite{Rakthanmanon11ICDM}. Practically, the introduction of support allows focusing on more frequent segments that occur at least $s$ times in a time series. Furthermore, it can contribute to obtain a more robust motif, as $s$ similar segments need to be combined to derive one motif.

In addition to the above-mentioned aspects, there are a number of known open issues in time series motif discovery, and in exact motif discovery in particular~\cite{Mueen14WIRES}. These include discovering repeated warped segments~\cite{Truong15BOOKCHAP}, finding motifs under complexity invariance~\cite{Batista11SDM} or uniform scaling~\cite{Yankov07KDD}, multi- and sub-dimensional motif discovery~\cite{Minnen07ICDM}, and algorithms that can work in anytime mode~\cite{Yingcharxxx13ICDM}. To the best of our knowledge, existing approaches address one or, at most, two of such open issues, with different tradeoffs and varying degrees of success. Moreover, they typically require to set a number of parameters. The only exception is SWARMMOTIF~\cite{Serra16KBS}, which adapts a specific metaheuristic~\cite{Luke13BOOK}, particle swarm optimization~\cite{Poli07SI}, to the task of time series motif discovery. Such adaptations can lead to very flexible algorithms that are able to tackle most of the previous issues at the same time~\cite{Serra16KBS}. We here consider an alternative metaheuristic adaptation that, in addition, goes beyond previous works by incorporating a novel notion of motif support in its formulation. 

In this paper, we present GENMOTIF, an approach that is based on the adaptation of genetic algorithms~\cite{Davis91BOOK} to the task of time series motif discovery. The main individual contributions of our approach are summarized below. However, we believe that it is the combination of such contributions what makes GENMOTIF a unique and novel algorithm for time series motif discovery:
\begin{itemize}
	\item It explicitly involves a new notion of motif support. As mentioned, this is a natural step that has both conceptual as well as practical benefits.
	\item It is an anytime algorithm. As with all genetic algorithms, an approximation to the final solution is incrementally refined with time.
	\item It does not oblige the practitioner to specify an exact segment length, which in many cases can be problematic.
	\item It performs uniform scaling of the segments that form a motif. This has been shown to be a desirable quality in a number of practical situations~\cite{Yankov07KDD}.
	\item It allows to analyze multi-dimensional time series.
	\item It is dissimilarity-agnostic. Its formulation is independent of the measurement used to assess segment similarity. This implies that we can measure segment similarity with, for example, dynamic time warping, or incorporate notions of complexity-invariance, if needed.
	\item It has only two parameters, which we show do not substantially affect its performance, provided that they are set within reasonable bounds.
\end{itemize}
%In addition, it should be noted that, by formulating GENMOTIF in terms of genetic algorithms, we also take advantage of all the flexibility that metaheuristics~\cite{Luke13BOOK} bring to the task of time series motif discovery~\cite{Serra15ARXIV}. Most prominently, GENMOTIF can directly work with elastic dissimilarity measures such and introduce the notion of complexity invariance. In future extensions, it could be also potentially adapted to work with sub-dimensional motifs.
We highlight the utility of GENMOTIF with one synthetic and three real-world case studies. 
%We highlight the utility of GENMOTIF with one synthetic and two real-world case studies. 
We also stress its novelty by performing a qualitative comparison with the most conceptually closest approaches in the literature. 
Except for one case study, we employ publicly-available data sets, and make all our code available online\footnote{\url{https://github.com/joansj/genmotif}} to foster reproducibility and to promote the usage of the algorithm in further scientific domains. 

The remainder of the paper is structured as follows. We first formulate the notion of motif support and introduce the necessary notation. We then detail the implementation of GENMOTIF. Next, we experimentally evaluate its performance and show two use cases that highlight its value. We also qualitatively compare GENMOTIF with the state-of-the-art. Finally, we summarize our contributions.

\section{PROBLEM DEFINITION AND NOTATION: MOTIFS WITH SUPPORT}
\label{sec:problemdef}

Consider a time series $\textbf{z}=[z_1,\dots z_n]$ of length $n$. Given a range of segment lengths of interest $l\in[l_{\min},l_{\max}]$, $l\ll n$, we want to find a set of $k$ motifs $\mathcal{M}=\lbrace \textbf{m}_1,\dots \textbf{m}_k\rbrace$ with a support of $s$ each. This means that a motif $\textbf{m}_i$ is derived from $s$ segments $\textbf{x}$ contained in $\textbf{z}$. A segment $\textbf{x}=[x_1,\dots x_l]$ is a small slice of $\textbf{z}$, that is, $[x_1,\dots x_l]=[z_{i+1},\dots z_{i+l}]$, $0\leq i\leq n-l$. In order to avoid trivial matches~\cite{Lin02WTDM}, the considered segments should preferably be non-overlapping, although some percent of overlap could be also tolerated~\cite{Mueen13ICDM}. Thus, a solution to the task is a set of $ks$ segments $\mathscr{X}=\lbrace \textbf{x}_1,\dots \textbf{x}_{ks} \rbrace$, where $k$ non-intersecting groups of $s$ segments each define a motif $\textbf{m}_i$. That is, $\mathscr{X}=\lbrace \mathcal{X}_1,\dots \mathcal{X}_k\rbrace$, $\mathcal{X}_i=\lbrace \textbf{x}_{u+1},\dots \textbf{x}_{u+s} \rbrace$, $u=s(i-1)$, $\mathcal{X}_i \cap \mathcal{X}_j = \emptyset$.

There are several ways in which a group of segments $\mathcal{X}_i$ can define a motif $\textbf{m}_i$, and these depend on the dissimilarity space and the central tendency measure chosen. Intuitively, $\textbf{m}_i$ should be a good representative of $\mathcal{X}_i$. Therefore, the motif $\textbf{m}_i$ should be `very close' to all $s$ segments in $\mathcal{X}_i$, according to a given dissimilarity measure. If we consider the Euclidean distance as the dissimilarity measure and the mean as a measure of central tendency, we can define
\begin{equation}
	\textbf{m}_i = \frac{1}{s} \sum_{\textbf{x}_u\subset \mathcal{X}_i} \textbf{x}_u .
	\label{eq:motif_average}
\end{equation}
The only requirement is that all segments $\textbf{x}_u$ are of equal length, which we can fulfill by simply upsampling to the largest length or by any other interpolation criterion. On a more general case, given an arbitrary dissimilarity measure between segments $\textbf{x}_u$ and $\textbf{x}_v$, $D(\textbf{x}_u,\textbf{x}_v)$, we can obtain $\textbf{m}_i$ by considering the medoid of $\mathcal{X}_i$, which we denote by $\textbf{x}_{u^\ast}$: 
\begin{equation}
	 \textbf{m}_i = \textbf{x}_{u^\ast} \,\,\mid\,\, u^\ast = \argmin\limits_{u \neq u^\ast} \left( \sum_{\textbf{x}_u\subset \mathcal{X}_i} D(\textbf{x}_u,\textbf{x}_{u^\ast}) \right) .
	 \label{eq:motif_medoid}
\end{equation}
It can be noted that with motif definitions in Eqs.~\ref{eq:motif_average} and~\ref{eq:motif_medoid}, there is an analogy between finding $k$ motifs with support and, respectively, $k$-means or $k$-medoids clustering~\cite{Xu09BOOK}. However, there is an important difference: we do not cluster all available data; only $ks$ short-length time series segments are considered. This can yield important speedups in terms of computation time, as we only consider a small part of all available segments, $ksl_{\max}\ll n$.

The previous definitions match our intuition that a motif $\textbf{m}_i$ should be `very close' to all segments in $\mathcal{X}_i$. However, there is another important factor that we may want to impose: the fact that motifs in $\mathcal{M}$ are maximally dissimilar between themselves. This will prevent the discovery of the same motif twice. To enforce the closeness of motifs' supports and, at the same time, the dissimilarity between different motifs, we can resort to existing clustering evaluation measures. For instance, we can consider the Davies-Bouldin index~\cite{Davies79PAMI} or the silhouette coefficient~\cite{Rousseeuw87JCAM}, which are both measures that take into account intra-cluster variance and inter-cluster dissimilarity~\cite{Xu09BOOK}. Following our notation, the Davies-Bouldin index corresponds to
\begin{equation}
	 I_{\text{DB}}(\mathscr{X}) = \frac{1}{k} \sum_{i=1}^{k} \max\limits_{\substack{1\leq j\leq k\\i\neq j}}\left( \frac{ R(\textbf{m}_i,\mathcal{X}_i) + R(\textbf{m}_j,\mathcal{X}_j) }{D(\textbf{m}_i,\textbf{m}_j)} \right) ,
	 \label{eq:dbi}
\end{equation}
and the silhouette index corresponds to
\begin{equation}
	I_{\text{S}}(\mathscr{X}) = 1-\frac{1}{ks} \sum_{u=1}^{ks} \frac{R(\textbf{x}_u,\mathcal{X}\setminus\mathcal{X}_i)-R(\textbf{x}_u,\mathcal{X}_i)}{\max(R(\textbf{x}_u,\mathcal{X}_i),R(\textbf{x}_u,\mathcal{X}\setminus\mathcal{X}_i))},
	\label{eq:silhouette}
\end{equation}
where
\begin{equation*}
	R(\textbf{y},\mathcal{X}) = \frac{1}{|\mathcal{X}|} \sum_{\textbf{x}_u\subset \mathcal{X}} D(\textbf{y},\textbf{x}_u)
\end{equation*}
and, in Eq.~\ref{eq:silhouette}, $\mathcal{X}_i$ denotes the set that contains $\textbf{x}_u$. Note that $I_{\text{DB}}$ makes explicit use of the concept of central tendency, while $I_{\text{S}}$ does not. Thus, depending on the dissimilarity measure used, one might be more appropriate than the other\footnote{Notice however that $I_{\text{S}}$ still requires some measure of central tendency to extract $\textbf{m}_i$ in a final stage.}.

Overall, finding $k$ motifs with $s$ support is a combinatorial problem. A solution consists of $ks$ segments, and if we have a time series of length $n$ and a range of segment lengths $r=l_{\max}+1-l_{\min}$, a brute-force search needs to examine, roughly, $\comb{nr}{ks}=\frac{(nr)!}{(ks)!(nr-ks)!}$ combinations. This, for a very small application involving $n=10\,000$, $r=100$, $k=2$, and $s=5$, already yields over $2.7\cdot 10^{53}$ possibilities\footnote{As a curiosity, we find there are about $1.33\cdot 10^{50}$ atoms in the world: \url{http://education.jlab.org/qa/mathatom_05.html}.}.

\section{GENMOTIF}
\label{sec:genmotif}

\subsection{Background}

In the following, we present our proposed approach, GENMOTIF, which is based on genetic algorithms. Genetic algorithms have a long tradition in solving combinatorial optimization problems~\cite{Davis91BOOK} as the one we face here. They are population-based metaheuristics~\cite{Luke13BOOK} that iterate an initial pool of solutions, improving a user-defined fitness criterion in every iteration. Genetic algorithms are known to yield good solutions with very few iterations, although it may take them much more to arrive at the global best solution possible. In some variants, such global best solution is proven to be reached, although no finite bound on the number of iterations can be established~\cite{Rudolph94TNN}. Thus, in essence, they can be seen as anytime algorithms~\cite{Zilberstein96AIM}. In a wider sense, metaheuristics have been recently introduced to the task of motif discovery, bringing in a high degree of flexibility with regards to the definition of the motif~\cite{Serra16KBS}. Such flexibility relies on the ability of metaheuristics to search complex, non-differentiable, discrete solution spaces under incomplete or imperfect information~\cite{Luke13BOOK}. 

\subsection{Solutions' mapping and fitness}

Before delving into the description of the approach, we first need to adapt the formulation of genetic algorithms to the specific problem of time series motif discovery. In genetic algorithms, a solution $\mathbb{X}$ has a set of properties, commonly called genes or genotypes, which can be combined and mutated. We propose to map $\mathscr{X}=\lbrace \textbf{x}_1,\dots \textbf{x}_{ks} \rbrace$ as a set of tuples $\mathbb{X}=\lbrace X_1,\dots X_{ks}\rbrace$ such that $X_i=\{f,l,c\}$, where $f$ is the index of the first sample of the time series segment, $l$ is the length of the segment, and $c$ is a real-valued indicator that will be used to determine the index of the motif to which the segment belongs to. Therefore, $f\in[1,n-l]$, $l\in[l_{\min},l_{\max}]$, and $c\in[0,1)$. Note that $X_i$ represents a time series segment regardless of its dimensionality, thus fitting it to multidimensional time series is directly applicable.

Once we have an encoding for our solutions $\mathbb{X}$, we need to specify the mapping from $\mathbb{X}$ to $\mathscr{X}$ and the criterion that will be used to evaluate the goodness of a solution $\mathscr{X}$ (Algorithm~\ref{alg:fitness}). We initiate the process by verifying that $\mathbb{X}$ is a valid solution (line~1). To do so, we loop over all the elements of $\mathbb{X}$ and check if they overlap with any other element in $\mathbb{X}$. In such case, we conclude that $\mathbb{X}$ is not a valid solution and return the value of $\infty$ (line~2), which is the worst possible value for a solution, as we will try to minimize $G$ (see below). If that is not the case, we proceed to copy the $ks$ segments defined by $\mathbb{X}$ to the array $\mathscr{X}$ and upsample them to $l_{\max}$ samples (line~3). For the latter we employ linear interpolation, but any other strategy would be applicable, depending on the domain. We next z-normalize each segment (line~4). Note that this step can be safely skipped if the motif definition  does not include such a constraint. After that, we get all motif group indicators contained in $\mathbb{X}$ (line~5): $\textbf{c}=\lbrace c_1,\dots c_{ks}\rbrace$, where $c_i$ corresponds to the third element of the tuple $X_i$. We then compute the indices $\textbf{o}$ that sort $\textbf{c}$ in increasing order (line~6). With $\textbf{o}$ and $s$, we can then group the segments in $\mathscr{X}$ so that each group has a direct correspondence to a motif (line~7). To do so, we take every $s$ consecutive indices in $\textbf{o}$ and form $k$ groups $\mathcal{X}_i$, $i=1,\dots k$. Specifically, $\mathcal{X}_i=\lbrace \textbf{x}_u,\dots \textbf{x}_v \rbrace$, where $u=o_{s(i-1)+1}$ and $v=o_{si}$. Finally, we compute the goodness of the solution (line~8) by applying a criterion of choice, as explained above. If not stated otherwise, we here use the Davies-Bouldin index (Eq.~\ref{eq:dbi}) for $I$, the squared Euclidean distance between z-normalized segments, and the mean as a measure of central tendency (Eq.~\ref{eq:motif_average}). Note that nothing prevents us from using another dissimilarity measurement or clustering index.

\begin{algorithm}[t]
	\caption{$G(\textbf{z},\mathbb{X},l_{\max},s)$}
	\label{alg:fitness}
	\begin{algorithmic}[1]
		\vspace*{0.1cm}
		\REQUIRE A time series $\textbf{z}$, a solution $\mathbb{X}$, the maximum length of the segments $l_{\max}$, and the motif support $s$. \\
		%{\renewcommand{\algorithmicrequire}{\textbf{Require:}} \REQUIRE A function $I$ assessing the goodness of a set of segments $\mathscr{Z}$ (see~\ref{sec:problemdef}). \\}
		\ENSURE A goodness value $g$ (the lower, the better). \\
		\vspace*{0.2cm}
		\IF{\textsc{SomeOverlap}$(\mathbb{X})$ }
			\RETURN $\infty$ \\
		\ENDIF
		\STATE $\mathscr{X} \leftarrow$ \textsc{CopyAndUpsampleSegments}$(\textbf{z},\mathbb{X},l_{\max})$ \\
		\STATE $\mathscr{X} \leftarrow$ \textsc{ZNormalizeSegments}$(\mathscr{X})$ \\
		\STATE $\textbf{c} \leftarrow$ \textsc{GetMotifIndicators}$(\mathbb{X})$ \\
		\STATE $\textbf{o} \leftarrow \text{arg}\,\text{sort}\,(\textbf{c})$ \\
		\STATE $\mathscr{X} \leftarrow$ \textsc{GroupSegments}$(\mathscr{X},\textbf{o},s)$ \\
		\RETURN $I(\mathscr{X})$ \\
	\end{algorithmic}
\end{algorithm}

\subsection{Main algorithm}

We are now ready to describe the main algorithm in GENMOTIF (Algorithm~\ref{alg:genetic}). GENMOTIF is an instance of a canonical genetic algorithm~\cite{Luke13BOOK}. However, it incorporates a basic form of elitism, performs a uniform crossover, and employs a convolutional mutation operator~\cite{Davis91BOOK}. In addition, it incorporates all the necessary adaptations to work with time series and time series segments. As an input, it takes a time series $\textbf{z}$, together with the minimum and maximum segment lengths of interest $l_{\min}$ and $l_{\max}$, the number of desired motifs $k$, and their support $s$. Its output represents the best solution $\mathbb{X}$ found in the available execution time $t_{\max}$. Note that all these variables are not parameters of the algorithm, but requirements of the task. With respect to the parameters, GENMOTIF requires only the population size $\rho$ and the mutation deviation constant $\sigma$ to be externally set. We will detail their use and study their effect below.

\begin{algorithm}[t]
	\caption{\textsc{GENMOTIF}$(\textbf{z},l_{\min},l_{\max},k,s,t_{\max})$}
	\label{alg:genetic}
	\begin{algorithmic}[1]
		\vspace*{0.1cm}
		\REQUIRE Time series $\textbf{z}$ of length $n$, minimum and maximum length of segments $l_{\min}$ and $l_{\max}$, number of motifs $k$, motif support $s$, and maximum execution time $t_{\max}$. \\
		{\renewcommand{\algorithmicrequire}{\textbf{Require:}} \REQUIRE Setting the size of the population $\rho$ and a mutation deviation constant $\sigma$. \\}
		\ENSURE Best solution found $\mathbb{X}$. \\
		\vspace*{0.2cm}
		%\STATE $t_0 \leftarrow$ \textsc{CurrentTime}$()$
		\STATE $\mathbb{P} \leftarrow \emptyset$ \\
		\FOR{$i=1,\dots \rho$}
			\STATE $\mathbb{X}_i \leftarrow$ \textsc{NewSolution}$(n,l_{\min},l_{\max},ks)$ \\
			\STATE $\mathbb{P} \leftarrow \mathbb{P} \cup \lbrace \mathbb{X}_i \rbrace$ \\
		\ENDFOR
		\LOOP
			\STATE $g_{\text{best}} \leftarrow \infty$  \\
			\FOR{$\mathbb{X}_i$ \textbf{in} $\mathbb{P}$}
				\STATE $g \leftarrow G(\textbf{z},\mathbb{X}_i,l_{\max},s)$ \\
				\IF{$g<g_{\text{best}}$}
					\STATE $g_{\text{best}} \leftarrow g$ \\
					\STATE $i_{\text{best}} \leftarrow i$ \\
				\ENDIF
			\ENDFOR
			%\IF{\textsc{CurrentTime}$()-t_0\geq t_{\max}$}
			\IF{\textsc{OutOfTime}$(t_{\max})$}
				\RETURN $\mathbb{X}_{i_{\text{best}}}$ \\
			\ENDIF
			\STATE $\mathbb{P}_{\text{new}} \leftarrow \lbrace \mathbb{X}_{i_{\text{best}}} \rbrace$ \\
			\FOR{$u=1,\dots (\rho-1)/2$}
				\STATE $\mathbb{X}_i,\mathbb{X}_j \leftarrow$ \textsc{ChooseAndCopy}$(\mathbb{P})$ \\
				\STATE $\mathbb{X}_i,\mathbb{X}_j \leftarrow $ \textsc{Crossover}$(\mathbb{X}_i,\mathbb{X}_j,ks)$ \\
				\STATE $\mathbb{X}_i \leftarrow$ \textsc{Mutation}$(\mathbb{X}_i,\sigma,n,l_{\min},l_{\max},ks)$ \\
				\STATE $\mathbb{X}_j \leftarrow$ \textsc{Mutation}$(\mathbb{X}_j,\sigma,n,l_{\min},l_{\max},ks)$ \\
				\STATE $\mathbb{P}_{\text{new}} \leftarrow \mathbb{P}_{\text{new}} \cup \lbrace \mathbb{X}_i,\mathbb{X}_j \rbrace$ \\
			\ENDFOR
			\STATE $\mathbb{P} \leftarrow \mathbb{P}_{\text{new}}$ \\
		\ENDLOOP
	\end{algorithmic}
\end{algorithm}

GENMOTIF starts by setting a new population of solutions $\mathbb{P}$ of size $\rho$, $\mathbb{P}=\lbrace \mathbb{X}_1,\dots \mathbb{X}_\rho\rbrace$, where $\rho$ is an odd integer such that $\rho\geq 3$ (Algorithm~\ref{alg:genetic}, lines~1--4). To instantiate a new solution $\mathbb{X}_i$ (line~3), we sample the search space uniformly at random. For $\mathbb{X}_i$ to be a valid solution, the sampling needs to be performed between reasonable limits, while avoiding potential segment overlaps (Algorithm~\ref{alg:newindividual}). To sample the search space we use a uniform real random number generator $U(a,b)\in [a,b)$ and the floor operator $\lfloor ~ \rfloor$. After computing the initial set of solutions $\mathbb{P}$, we enter an infinite loop (Algorithm~\ref{alg:genetic}, lines~5--21). This is the loop that will evolve the solutions in the population. It performs three main operations: (a) selecting the best fit solution, (b) checking execution time, and (c) constructing a new population. 

For selecting the best solution (lines~6--11), we just compute the goodness value $G$ as explained before (line~8; see Algorithm~\ref{alg:fitness}) and keep the index $i_{\text{best}}$ of the solution that has the minimum value (line~11). Once all $\rho$ solutions have been examined, we check the set time constraint $t_{\max}$ (lines~12--13). If the remaining time does not permit the new iteration, we return the best individual found until that point, $\mathbb{X}_{i_{\text{best}}}$ (line~13). Alternatively, if there is time left, we continue by constructing a new population $\mathbb{P}_{\text{new}}$ (lines~14--20) and replacing $\mathbb{P}$ with this new population (line~21). Importantly, $\mathbb{P}_{\text{new}}$ keeps the best solution found at every iteration, unaltered (line~14). This is the simplest form of elitism possible~\cite{Luke13BOOK}, and the minimum requirement for the global convergence of the algorithm~\cite{Rudolph94TNN}. The construction of $\mathbb{P}_{\text{new}}$ continues by uniformly sampling two individuals $\mathbb{X}_i,\mathbb{X}_j$ with replacement from $\mathbb{P}$, and returning a copy of them (line~16). Next, we cross their genes by uniform crossover (line~17) and mutate their genes according to a mutation factor $\sigma$ (lines~18--19).  Finally, $\mathbb{X}_i$ and $\mathbb{X}_j$ are appended to $\mathbb{P}_{\text{new}}$ (line~20).

\begin{algorithm}[t]
	\caption{\textsc{NewSolution}$(n,l_{\min},l_{\max},ks)$}
	\label{alg:newindividual}
	\begin{algorithmic}[1]
		\vspace*{0.1cm}
		\REQUIRE Length of the series $n$, minimum and maximum segment lengths $l_{\min}$ and $l_{\max}$, and number of segments $ks$. \\
		%{\renewcommand{\algorithmicrequire}{\textbf{Require:}} \REQUIRE A function $I$ assessing the goodness of a set of segments $\mathscr{Z}$ (see~\ref{sec:problemdef}). \\}
		\ENSURE A valid solution $\mathbb{X}$. \\
		\vspace*{0.2cm}
		\STATE $\mathbb{X} \leftarrow \emptyset$ \\
		\WHILE{\textsc{IsEmpty}$(\mathbb{X})$ \textbf{or} \textsc{SomeOverlap}$(\mathbb{X})$}
			\STATE $\mathbb{X} \leftarrow \emptyset$ \\
			\FOR{$i=1,\dots ks$}
				\STATE $c \leftarrow U(0,1)$ \\
				\STATE $l \leftarrow \lfloor U(l_{\min},l_{\max}+1) \rfloor$ \\
				\STATE $f \leftarrow \lfloor U(1,n-l+1) \rfloor$ \\
				\STATE $\mathbb{X} \leftarrow \mathbb{X} \cup \lbrace f,l,c \rbrace$ \\
			\ENDFOR
		\ENDWHILE
		\RETURN $\mathbb{X}$ \\
	\end{algorithmic}
\end{algorithm}

The crossover and mutation operators require a careful design consideration. In GENMOTIF, we opted for a uniform crossover with a fixed low probability (Algorithm~\ref{alg:crossover}). With probability $1/ks$ (line~2), we swap the $u$-th segments of the two solutions (line~3). Similarly, for the mutation operator, with probability $1/ks$ (line~3), we mutate a segment $u$ (lines~4--8). To do so, we follow a convolutional process (Algorithm~\ref{alg:mutation}). At the end of the mutation operation, we shuffle the segments in $\mathbb{X}$ to favor the exploration of the search space (line~9). In pre-analysis, we found this shuffling operation to have a marginal but positive impact to the fitness of the found solutions, in particular for short execution times.

\begin{algorithm}[t]
	\caption{\textsc{Crossover}$(\mathbb{X}_i,\mathbb{X}_j,ks)$}
	\label{alg:crossover}
	\begin{algorithmic}[1]
		\vspace*{0.1cm}
		\REQUIRE Two solutions $\mathbb{X}_i,\mathbb{X}_j$ and number of segments $ks$. \\
		\ENSURE Solutions $\mathbb{X}_i$ and $\mathbb{X}_j$ with some genes crossed. \\
		\vspace*{0.2cm}
		\FOR{$X_{i_u},X_{j_u}$ \textbf{in} $\mathbb{X}_i,\mathbb{X}_j$}
			\IF{$U(0,1)<1/ks$}
				\STATE \textsc{Swap}$(X_{i_u},X_{j_u})$ \\
			\ENDIF
		\ENDFOR
		\RETURN $\mathbb{X}_i,\mathbb{X}_j$ \\
	\end{algorithmic}
\end{algorithm}

\begin{algorithm}[t]
	\caption{\textsc{Mutation}$(\mathbb{X},\sigma,n,l_{\min},l_{\max},ks)$}
	\label{alg:mutation}
	\begin{algorithmic}[1]
		\vspace*{0.1cm}
		\REQUIRE Solution $\mathbb{X}$, mutation parameter $\sigma$, length of the time series $n$,  minimum and maximum segment length $l_{\min}$ and $l_{\max}$, and number of segments $ks$. \\
		\ENSURE A mutation of the solution $\mathbb{X}$. \\
		\vspace*{0.2cm}
		\STATE $r \leftarrow l_{\max}+1-l_{\min}$ \\
		\FOR{$X_u$ \textbf{in} $\mathbb{X}$}
			\IF{$U(0,1)<1/ks$}
				\STATE $f,l,c \leftarrow X_u$ \\
				\STATE $c \leftarrow \text{mod}\left(c+\sigma C(0,1),1\right)$ \\
				\STATE $l \leftarrow \text{mod}\left(l+\lfloor \sigma rC(0,1)\rceil -l_{\min},r+1\right) +l_{\min}$ \\
				\STATE $f \leftarrow \text{mod}\left(f+\lfloor \sigma (n-l)C(0,1)\rceil,n-l+1 \right)$ \\
				\STATE $X_u \leftarrow \lbrace f,l,c \rbrace$\\
			\ENDIF
		\ENDFOR
		\STATE \textsc{Shuffle}$(\mathbb{X})$ \\
		\RETURN $\mathbb{X}$ \\
	\end{algorithmic}
\end{algorithm}

To mutate the values $f$, $l$, and $c$, we employ a real random number generator following the standard Cauchy distribution $C(0,1)$, which has the location and scale parameters set to 0 and 1, respectively. The Cauchy distribution is a heavy-tailed distribution, with undefined mean and variance. It can be considered to be a special case of the more general L\'evy flight distribution. L\'evy flight distributions have been suggested to be beneficial for optimization-based metaheuristics when searching large spaces, as very long jumps can be performed after many small ones~\cite{Yang10BOOK}. Moreover, a L\'evy flight strategy is reported to be an efficient statistical strategy for searching randomly located, sparse objects~\cite{Viswanathan99NATURE}. In the absence of a~priori knowledge about the distribution of the targets, the case corresponding to the Cauchy distribution has been shown to be the optimal search strategy~\cite{Viswanathan99NATURE}. 

To compute $C(0,1)$, we just divide two real random numbers taken from the normal distribution with zero mean and unit variance, $N(0,1)$, such that $C(0,1) = N(0,1)/\left| N(0,1) \right|$, where $\left| ~\right|$ indicates absolute value. We then rescale $C(0,1)$ by the ranges of the corresponding variables (Algorithm~\ref{alg:mutation}, lines~5--7) and control the mutation variability by the parameter $\sigma$. We additionally employ the round to the nearest integer operator $\lfloor ~\rceil$ where needed and control the limits of the mutated variables with the modulo operator $\text{mod}()$.

\section{EXPERIMENTAL VALIDATION}
\label{sec:evaluation}

In this section, we evaluate GENMOTIF experimentally. We start with a sanity check by a planted motif problem. We then study the influence of the two parameters of the algorithm. We run all our experiments on a 4~GB RAM laptop with a dual core processor at 2.5~GHz. The only exception are the parameter setting experiments, which were run on a 15~GB RAM machine with 24 cores at 2~GHz. Except for the planted motif time series, which are artificially generated, all data sets used in this section are publicly-available.

\subsection{Planted motifs}

\begin{figure}[t]
	\centering
	\includegraphics{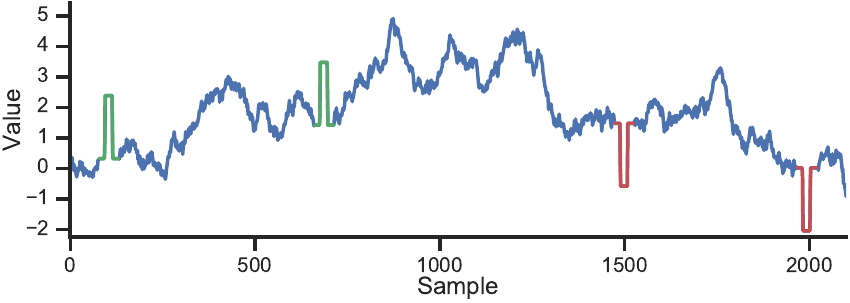}
	\caption{First samples of the random walk time series with planted square wave motifs. In this and subsequent time series plots, horizontal and vertical axes correspond to samples and values, respectively.}
	\label{fig:planted_ts}
\end{figure}

We first consider a planted motif task where motifs have been downsampled uniformly at random to, at most, 90\% of their original length. We insert two different patterns corresponding to a square wave of length 58 in a random walk time series generated by $x_i=x_{i-1}+N(0,0.1)$ with $x_0=0$ (Fig.~\ref{fig:planted_ts}). We arbitrarily consider 50 downsampled repetitions for each pattern and force that they represent only 10\% of the total length of the time series, which results in 58\,000~samples. We then run GENMOTIF to find 4 motifs with a support of 5 and lengths between 50 and 60 samples.

\begin{figure}[t]
	\centering
	\includegraphics{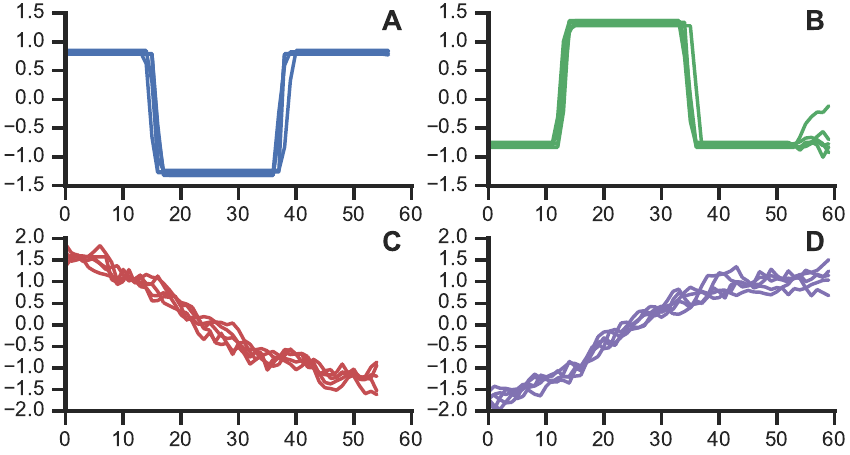}
	\caption{Solution found in the planted motif random walk time series after a two minutes run. Motifs A and B correspond to the planted motifs, while motifs C and D are two motifs that are expected to be found in a sufficiently long random walk time series.}
	\label{fig:planted_sol}
\end{figure}

The results after two minutes computation time\footnote{We run all experiments on a single core at 2.5~GHz except the parameter setting experiments, using single cores of 2~GHz.} are already satisfactory (Fig.~\ref{fig:planted_sol}). The algorithm is able to discover the two planted motifs (motifs A and B) and also two clear patterns that are trivially present in the random walk time series (motifs C and D). The lengths for the supports of motifs A and B are found to be between 52 and 60 samples. However, upsampling to the largest length found in the motif supports results in motif A having a length of 58 and motif B having a length of 60, the former being the original motif length and the latter being very close to it. We can also witness that motif B is not already centered to its most optimal value. This is a typical outcome when running time is relatively short and which improves as time progresses\footnote{It should be mentioned that, with an execution time of 1\,min, such motifs were already visible, although with a different centering and with less defined supports.}. Note, however, that this fact does not prevent the practitioner to `discover' the two originally planted motifs.

\begin{figure}[t]
	\centering
	\includegraphics{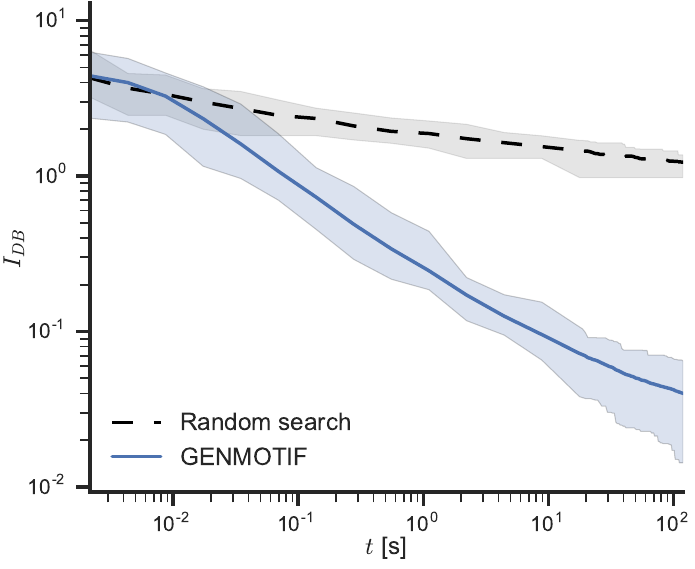}
	\caption{Solution fitness $I_{\text{DB}}$ as a function of execution time $t$. The dashed black line corresponds to a random search strategy and the solid blue line corresponds to GENMOTIF. The shaded regions correspond to 95\% prediction intervals. Notice the double logarithmic axes.}
	\label{fig:planted_fit}
\end{figure}

Given the anytime nature of the proposed algorithm, it is worth studying its convergence as execution time progresses. To do so, we analyze the goodness of fit index $I_{\text{DB}}$ as a function of the execution time $t$ (Fig.~\ref{fig:planted_fit}). In addition, we compare it with the $I_{\text{DB}}$ obtained by a simple random search strategy that uniformly samples the search space at every iteration. We see that both GENMOTIF and random search strategies improve the solution fitness with $t$. However, GENMOTIF performs considerably faster (notice the double logarithmic axes). Specifically, we see that the prediction intervals of the solutions found by the two approaches do not overlap after 100\,ms. This difference steadily increases with $t$ and, after 100\,s, it goes up to two orders of magnitude (Fig.~\ref{fig:planted_fit}).

\subsection{Parameter setting}
\label{sec:experimental_params}

\begin{figure}[t]
	\centering
	\includegraphics{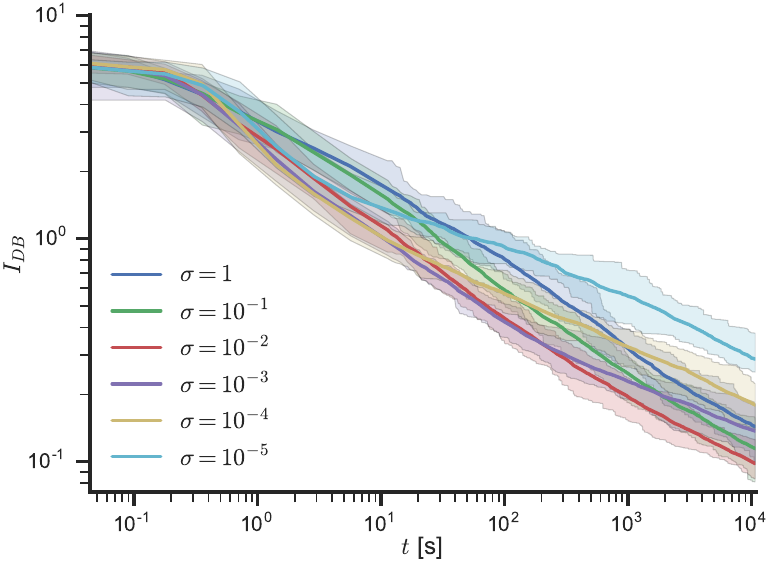}
	\caption{Solution fitness $I_{\text{DB}}$ as a function of execution time $t$, considering different values of $\sigma$ and using the EEG time series from~\cite{Mueen09SDM}. The shaded regions correspond to 95\% prediction intervals. Notice the double logarithmic axes.}
	\label{fig:param_sigma}
\end{figure}

In the following, we study the impact that the population size $\rho$ and the mutation deviation constant $\sigma$ make on the goodness of the solutions obtained by GENMOTIF. We start by setting $\rho=51$ and computing $I_{\text{DB}}$ as a function of time for $\sigma=\lbrace 10^{-5},10^{-4},10^{-3},10^{-2},10^{-1},1 \rbrace$ (Fig.~\ref{fig:param_sigma}). We first consider the EEG time series from~\cite{Mueen09SDM} and search for 10 motifs with a support of 5 with lengths between 150 and 250 samples. We see that the prediction intervals for the majority of the considered $\sigma$ values highly overlap, with $\sigma\in[10^{-1},10^{-3}]$ yielding only slightly better results than $\sigma$ values outside this range. We also see that $\sigma=10^{-5}$ yields the worst performance, with a prediction interval that does not overlap with the ones of the other $\sigma$ values obtained for approximately $t>5$\,min.

We repeated the experiment with the 9 time series and motif lengths considered in~\cite{Serra16KBS}, and obtained similar results to the ones of the previous paragraph. In general, a value of $\sigma=10^{-2}$ is expected to yield the best performance in the big majority of the cases, with $\sigma=10^{-3}$ typically performing almost as good as $\sigma=10^{-2}$. Interestingly, the value of $\sigma=10^{-2}$ is the default scaling value used in other studies employing L\'evy flight distributions for alternative metaheuristic algorithms~\cite{Yang10BOOK}. Thus, the $\sigma$ value we find in our experiments is consistent with current practice. This suggests a potentially optimal setting for these type of distributions that could be further investigated in future works and optimization scenarios.

\begin{figure}[t]
	\centering
	\includegraphics{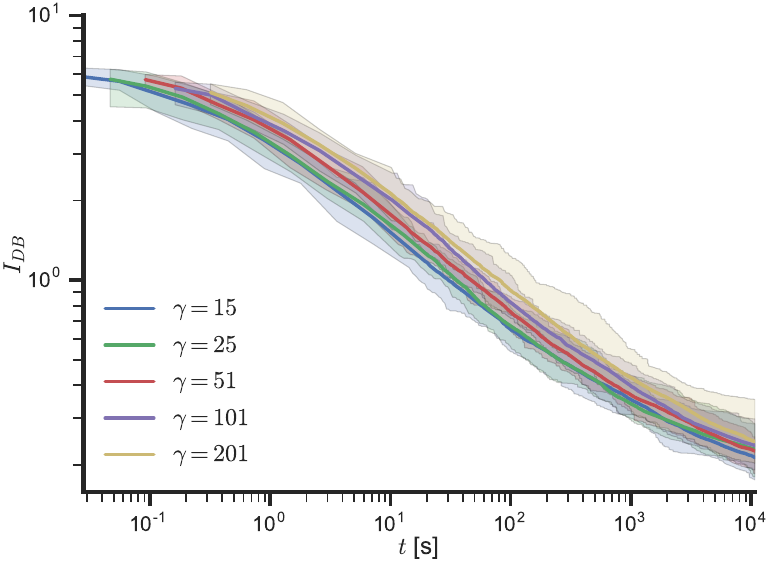}
	\caption{Solution fitness $I_{\text{DB}}$ as a function of execution time $t$, considering different values of $\rho$ and using the long Insect time series from~\cite{Mueen09SDM}. The shaded regions correspond to 95\% prediction intervals. Notice the double logarithmic axes.}
	\label{fig:param_popsize}
\end{figure}

We now focus on the analysis of the population size $\rho$. We set $\sigma=10^{-2}$ and compute $I_{\text{DB}}$ as a function of time for $\rho=\lbrace 15,25,51,101,201\rbrace$ (Fig.~\ref{fig:param_popsize}). We first consider the long Insect time series from~\cite{Mueen09SDM} and search for 10 motifs with a support of 5 with lengths between 300 and 500 samples. We see that, in this case, the overlap between prediction intervals becomes considerable. This suggests that $\rho$ has less overall effect than $\sigma$, and therefore that its setting can be less precise. 

We again repeated the experiment with the 9 time series and motif lengths considered in~\cite{Serra16KBS}, and obtained qualitatively similar results to the ones of the previous paragraph. We witnessed a minor tendency towards small population sizes (e.g.,~15 or 25 individuals) for short running times (e.g.,~1\,hour) and larger population sizes (e.g.,~51 or 101 individuals) for longer running times. Such an outcome is theoretically plausible, considering the fact that the larger the search space and the available time, the larger the exploration capabilities that a genetic algorithm requires; and increasing the population size is an elementary way of increasing exploration~\cite{Davis91BOOK}. However, as prediction intervals were found to be highly overlapping, we are not in a position to draw any strong statement in this regard.

\section{CASE STUDIES}
\label{sec:case_studies}

In this section, we demonstrate the usage of GENMOTIF in three different real-world scenarios that have already attracted the attention of the scientific community. 
%In this section, we demonstrate the usage of GENMOTIF in two real-world scenarios that have already attracted the attention of the scientific community. 
The case studies shown here are not designed to claim results in their respective  domains. Instead, they illustrate the applicability and the values of GENMOTIF in such domains. 
We consider tree types of signals: traffic volume, accelerometer, and telephone call data. The first two data sets are publicly-available while the third one is not open to the public for data proprietary reasons.
%We consider two types of signals: traffic volume and accelerometer data. Both data sets are publicly-available.

\subsection{Traffic volume patterns}

Traffic patterns and mobility of people in general is an important variable to consider in epidemics spread modeling, road dimensioning, and path optimization, to name a few. Therefore, timely modeling of traffic patterns is an area of paramount importance for both public and private sectors. To illustrate GENMOTIF's performance and versatility, we use the vehicle counts measured every 5\,min using an on-ramp sensor on the 101 freeway in Los Angeles, located near the Dodger Stadium, where the Los Angeles Dodgers baseball team plays their home games~\cite{Ihler06KDD}. Our goal is to evaluate how well GENMOTIF is able to identify objectively-relevant motifs in a real-world time series of 50\,400~samples.

\begin{figure}[t]
	\centering
	\includegraphics{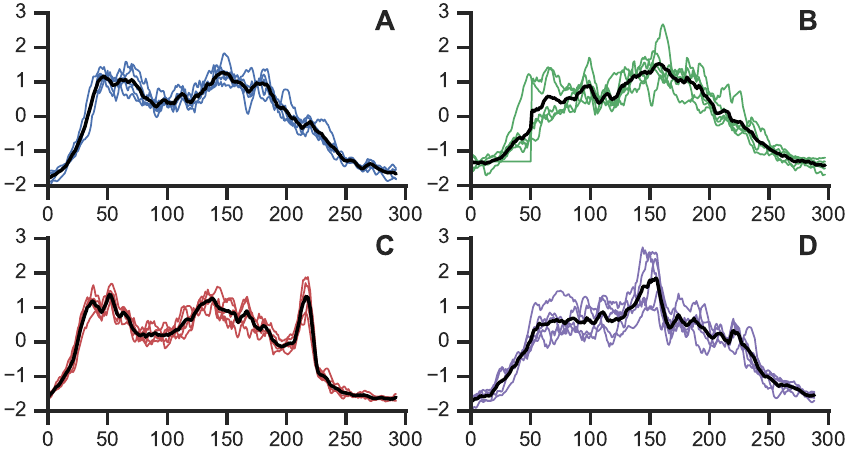}
	\caption{Four motifs found in the traffic volume data used in~\cite{Ihler06KDD} by GENMOTIF. Motifs A and B correspond to the typical traffic volume patterns of a work day and a weekend day, respectively. Motifs C and D correspond to the same situation, respectively, but when there is a match taking place. The solid black curve depicts the average of the motif supports.}
	\label{fig:case_traffic}
\end{figure}

We use the data available in the UCI archive\footnote{\url{https://archive.ics.uci.edu/ml/datasets/Dodgers+Loop+Sensor}}~\cite{Ihler06KDD} and run GENMOTIF for 10\,min, searching for 10 motifs with a support of 5 and lengths between 276 and 300 samples (corresponding to 23 and 25\,hours, respectively). Among the identified motifs, four patterns clearly stand out (Fig.~\ref{fig:case_traffic}). By knowing the context of the patterns (the date and the stadium events), these four patterns have straightforward and meaningful interpretations. Firstly, it is possible to clearly differentiate between working day patterns (motifs A and C) and weekend day patterns (motifs B and D). Moreover, these patterns additionally differ depending on whether the Dodgers played in their stadium (motifs C and D) or not (motifs A and B). The UCI archive documentation further contributes to understanding the relevance of this result, as it states that the sensor ``is close enough to the stadium to see unusual traffic after a Dodgers game, but not so close and heavily used by game traffic so that the signal for the extra traffic is overly obvious''.

\subsection{Person verification using accelerometer data}

Finding patterns in accelerometer data is an important research area with a wide applicability that spans from user interaction and activity tracking to health care assessment and biometric authentication. The accelerometer signal is a time series with a sampling rate that varies with the application: from 10--20\,Hz for phone interaction applications to, for instance, 500--1000\,Hz for detecting anomalies in engines. In our example, we apply GENMOTIF to the data captured from a single chest accelerometer with the objective of person verification~\cite{Casale10PUC}. Our goal is to find characteristic patterns for different persons in a time series of near 2\,million samples that features 52\,Hz recordings of the daily activities of 15 people including walking, talking, or working on a computer. 

\begin{figure}[t]
	\centering
	\includegraphics{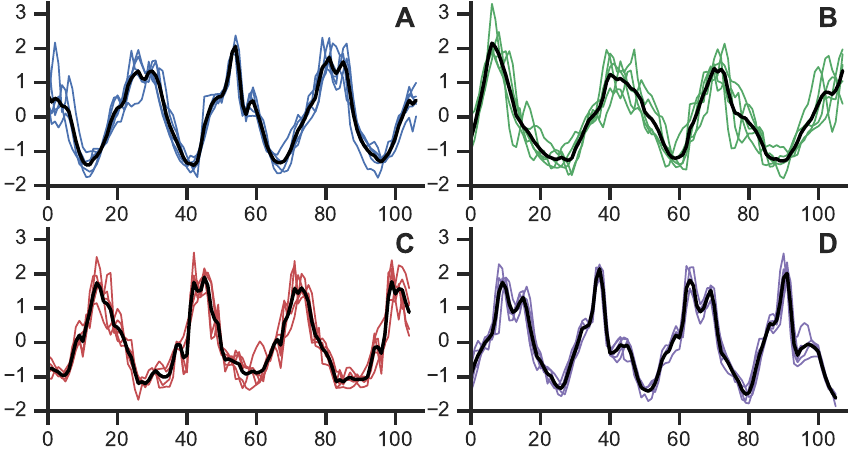}
	\caption{Four motifs found in the single chest accelerometer data from~\cite{Casale10PUC}. Motifs A, B, C, and D correspond to four different people. The solid black curve depicts the average of the motif supports.}
	\label{fig:case_person}
\end{figure}

We use the data available in the UCI archive\footnote{\url{https://archive.ics.uci.edu/ml/datasets/Activity+Recognition+from+Single+Chest-Mounted+Accelerometer}}~\cite{Casale10PUC}. We compute the modulus of the three acceleration coordinates and concatenate all signals from the 15 participants in a single time series. We then run GENMOTIF for a maximum execution time of over an hour, looking for 20 motifs with a support of 5 and lengths between 100 and 108 samples (around 1\,s). Among the motifs found, we could clearly observe a number of person-characteristic patterns (Fig.~\ref{fig:case_person}). Interestingly, many of them corresponded to walking patterns, which is the main activity considered in previous works for person verification~\cite{Casale10PUC}.

After that, we studied the correspondence between the obtained motifs and the labels provided in the data set (Table~\ref{tab:case_person}). The rationale is that, if motif supports mostly belong to a unique person, we can hypothesize that the inferred patterns are a characteristic of that person (in other words, his/her `walking fingerprint'). For each group of segments supporting a motif, we computed the mode with respect to the person identifier and, if there was a clear `winner' (that is, if a person had clearly more supports than any other), we assigned that motif to the corresponding person. Depending on the algorithm's running time, we were able to reliably identify patterns from 8 to 12 out of the 15 persons in the data set (Table~\ref{tab:case_person}, second column). Expectedly, the number of identified person-assigned patterns does not linearly depend on the execution time: after 8\,min, the number of different persons with assigned motifs already reaches 11, which roughly corresponds to a 73\% recall, and after 64\,min it reaches 12, which corresponds to an 80\% recall. Interestingly, as execution time progresses, we also see that not only more person-related motifs appear, but also that the motif supports are increasingly becoming associated to a single person (Table~\ref{tab:case_person}, third column) and to a single task (Table~\ref{tab:case_person}, fourth column). This shows that the refinements made by GENMOTIF to the solutions found for this task go towards the direction of having single-person and single-task support-based motifs.

\begin{table}
\centering
\caption{Comparing the information of the obtained motifs and supports with the ground truth given for the data set of~\cite{Casale10PUC}.}
\resizebox{1\linewidth}{!}{
\begin{tabular}{c|ccc}
\hline\hline
Runtime 	& Persons with	& Person dominance	& Task dominance \\
(min)   	& motif 		& in motif 			& in motif \\
\hline
1			& 8/15			& 51.8\%			& 57.6\%	\\
2			& 9/15			& 57.9\%			& 68.4\%	\\
4			& 10/15			& 68.0\%			& 70.7\%	\\
8			& 11/15			& 67.9\%			& 68.9\%	\\
16			& 11/15			& 74.1\%			& 77.6\%	\\
32			& 11/15			& 76.7\%			& 86.7\%	\\
64			& 12/15			& 85.3\%			& 89.3\%	\\
\hline\hline
\end{tabular}
}
\label{tab:case_person}
\end{table}

\subsection{Exploring personalities with outgoing call motifs}

Human communication patterns have long attracted the attention of multiple scientific disciplines. Nowadays, a considerable portion of our communications leave digital traces (such as phone calls, emails, short messages, etc.), a fact that provides the historical opportunity to quantify these interactions on a large-scale and to advance the theoretical knowledge about human interaction dynamics. In particular, patterns of mobile phone communications (that is, voice calls and short messages) have been used to study social behavior, to distinguish between different user profiles (such as ordinary users versus sales persons, telecom frauds, and robot-based callers), and even to infer the users' personality~\cite{Oliveira11CHI}. In this case study, we explore the applicability of GENMOTIF to discover patterns in time series of mobile phone calls. Specifically, we are interested in sequences of total daily outgoing call durations with lengths around 1\ month. 
%The data was gathered in a study described in~\cite{Oliveira13TOCHI}, and it involved approximately 600 participants who filled out a personality questionnaire and gave their consent to use anonymized call detailed records. %It should be mentioned that this use case was discussed on a speculative level and we refrain from any strong conclusions. 
To generate the time series, we use an internal data set of anonymized call detail records of approximately 600 subjects who, in addition, kindly filled out the Big 5 personality questionnaire\footnote{\url{http://ipip.ori.org/New_IPIP-50-item-scale.htm}}~\cite{Goldberg92PS}. We compute the sum of daily outgoing call durations for a total of 6 months per subject, and concatenate these into a single time series of over 108\,000 samples.

\begin{figure}[t]
	\includegraphics{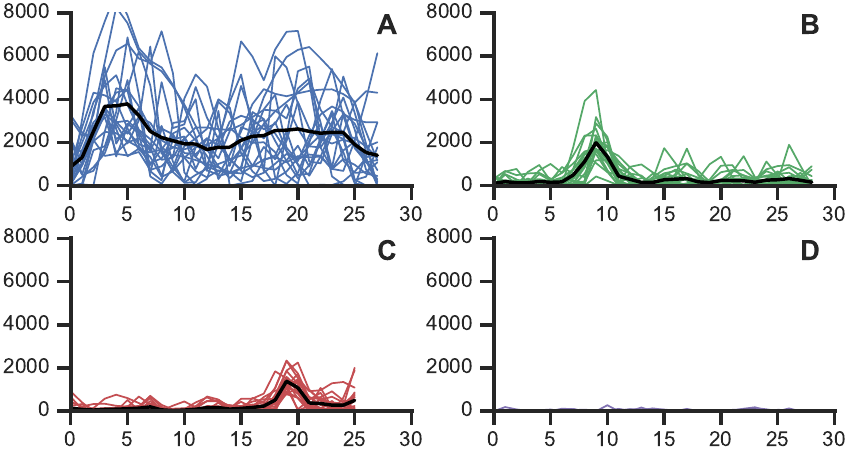}
	\caption{Four motifs found in the telephone calls time series by GENMOTIF. The vertical axis corresponds to the sum of call durations (in s) and the horizontal axis corresponds to days. The solid black curve depicts the average of the motif supports. In this case, no z-normalization was used.}
	\label{fig:case_calls}
\end{figure}

Four different patterns were identified after running GENMOTIF for 5 minutes (Fig.~\ref{fig:case_calls}). To obtain them, we searched for 5 motifs with a support of 20 and lengths between 14 and 31 samples (corresponding to 2 weeks and 1 month, respectively). We further analyzed how these patterns corresponded to different personalities of the participants (as reported in the questionnaire). It turned out that motif A predominantly corresponded to users with low conscientious scores (70\% of them scored lower than a mean value of the involved sample), whereas motif C corresponded to participants who scored above the average. We can speculate that motif A reflects a more random pattern of calls than motif C, thus reflecting less conscientious profiles. Similarly, motifs B and D distinguished participants based on the extraversion component of the questionnaire (again with 70\%). Motif B corresponded to the participants with the scores above the average and motif D was associated to the participants who scored below the average. Again, one may speculate that the volume of outgoing calls reflects the level of initiated social interactions that further mirrors the extraverted personality. Interestingly, motifs B and C are similar with respect to a peak that appears once in one month, yet motif B clustered mainly extraverted individuals (without any substantial difference in the proportion of conscientious scores) whereas the motif C mainly grouped individuals whose conscientiousness score was lower than the average (without any substantial difference in the proportion of extraversion scores in this pattern). This may indicate that the shape of the peak was a distinguishing factor: the higher amount of durations in motif B might indicate the extraverted nature of participants, whereas highly conscientious individuals have a more regular baseline of their communication patterns (perceived in motif C). Note that in previous studies related to inferring personalities through the call patterns (e.g.,~\cite{Montjoye13BOOKCHAP,Oliveira11CHI}), several hundreds of features were extracted to model the five components of personality, as opposed to only one signal, namely daily call durations, that we used in this example.

\section{RELATED WORK}
\label{sec:soa}

The approach that is the most similar to GENMOTIF is the one proposed in~\cite{Serra16KBS}. In such work, particle swarm optimization~\cite{Poli07SI} is used to find pairs of motifs under uniform scaling in an anytime fashion. The approach can also find warped motifs and it can work with multidimensional time series. However, the approach does not consider the notion of motifs with support nor does it guarantee that the found motifs are dissimilar between themselves. Both features are key contributions introduced by GENMOTIF. %Besides, the authors of~\cite{Serra15ARXIV} did not specifically consider any case study where their approach could be directly applied. We here have made a strong effort to highlight relevant applications of GENMOTIF.

In~\cite{Yingcharxxx13ICDM}, an anytime algorithm to find motifs in time series is presented. It uses the minimum description length principle as a criterion to stop its execution as well as a measure of segment similarity. This algorithm does not consider motif invariances such as warpings or uniform scaling, and its application to multidimensional time series was not considered. In~\cite{Mueen13ICDM}, an efficient algorithm for enumerating motifs in a range of lengths is explained. This is not an anytime algorithm nor does it involve the notion of motif supports. The same happens with~\cite{Yankov07KDD}, which considers the task of finding motifs under uniform scaling and exploits an approximate symbolic representation. The pioneering algorithm proposed in~\cite{Mueen09SDM} searches for motif pairs of a specific length, and its adaptation to a range of lengths (made by the same authors) has been shown to underperform the approach in~\cite{Mueen13ICDM}. There other countless approaches that could be related to GENMOTIF, including approximate motif discovery algorithms~\cite{Begum14VLDB,Castro10SDM,Chiu03KDD,Li12SDM}. However, one can easily find some task dimension where they significantly differ from GENMOTIF.

Other algorithms have been proposed for the task of time series subsequence clustering~\cite{Madicar13KST,Rakthanmanon11ICDM}. Though this task and the one of finding motifs with support have common aspects, they are conceptually different. Time series subsequence clustering aims at obtaining a good and compact representation of the whole time series while, as shown in~\cite{Rakthanmanon11ICDM}, ignoring some data. However, motif discovery aims at finding characteristic patterns (with support) that do not necessarily need to reliably and compactly represent the full time series. In fact, the segments used to derive the motifs can actually represent a tiny part of the time series (i.e., $ksl_{\max}\ll n$, as mentioned in the main text). 

Finally, we should note that the concept of motifs' support has been formulated in the past by some authors in a different way, specifically involving the choice of a dissimilarity threshold~\cite{Lin02WTDM,Mueen14WIRES}. Essentially, such formulations reduce the task of motif discovery to what we could call a `bucket-based' or a `radius-based' approach. For instance, one can think of using a locality-sensitive hashing algorithm which, after selecting an appropriate threshold, could distribute all possible segments of a time series into different buckets. Consequently, all segments in a bucket would determine a motif, and the support of such motif would be the number of segments in the bucket. This type of approaches face several limitations. Firstly, a single threshold is used for all motifs. This implies the assumption that all motifs form `clusters' of the same similarity radius, which is not necessarily the case for real-world data sets. Secondly, dissimilarity thresholds can be extremely difficult to estimate. As shown elsewhere~\cite{Chavez01ACMCS}, the variability of the distances between the elements in a metric space substantially reduces due to the curse of dimensionality. Therefore, with increasing dimensionality, setting such threshold becomes more problematic. Thirdly, different segment lengths require specifically tailored solutions. In this regard, if one aims at discovering motifs within a range of lengths, the aforementioned `radius-based' approach is not intuitive anymore as in the case of a fixed length. Lastly, it becomes challenging to include uniform scaling or other invariances in the definition of motif. Overall, we believe that the threshold-based definition of motif support is much more restrictive than the notion considered here.

%Developing any post-processing on top of an existing motif-finding algorithm could also fall into the aforementioned limitations. Moreover, such post-processing would have to handle potentially missing segments due to overlap (or handle overlapping segments), and it would have to be very efficient, as the number of candidates could be much larger than the size of the raw time series. This, in turn, could affect the execution time of the original motif discovery algorithm, specially of those based on lower-bounding. \comm{all}{Should I remove this paragraph? It is a little bit vague...}

\section{CONCLUSION}
\label{sec:conclusion}

In this study, we tackled the problem of finding motifs with support. We proposed GENMOTIF, a genetic algorithm that solves the problem in an anytime fashion and that, at the same time, is flexible enough to accommodate multiple definitions of motif. As it is solely based on dissimilarity measurements and clustering criteria, GENMOTIF can work, for instance, with a range of lengths, uniformly-scaled motifs, warped or complexity-invariant similarity measurements, and multidimensional time series. 

We demonstrated the utility of GENMOTIF in a number of tasks, including synthetic data, traffic volume measurements, accelerometer signals, and telephone call records. 
%We demonstrated the utility of GENMOTIF in a number of tasks, including synthetic data, traffic volume measurements, and accelerometer signals. 
We have also shown that the two parameters of the algorithm do not critically affect the performance of GENMOTIF, and we have provided a quantitative as well as conceptual discussion in this regard. 
In all our experiments, we have used publicly-available data, except for one case study where data was not open to the public due to proprietary reasons. 
%In all our experiments, we have used publicly-available data. 
Finally, we made GENMOTIF's code available online in order to foster research in time series motif discovery algorithms and to encourage the use of these algorithms in further scientific domains.

\section*{Acknowledgment}

We would like to thank all the donors of the data sets used in this study. This research has been partially funded by ICT-2014-15-645323 from the European Commission (JS) and 2014-SGR-118 from Generalitat de Catalunya (JLA).

\bibliographystyle{IEEEtran}
\bibliography{IEEEabrv,joan}

\end{document}